\documentclass[a4paper]{./styles/svproc}
\usepackage{url}
\usepackage{xcolor}
\usepackage{amsfonts}
\usepackage{hyperref}
\usepackage{graphicx}
\usepackage[style=ieee,backend=biber]{biblatex}
\usepackage{verbatim}

\AtEveryBibitem{%
  \clearfield{volume}%
  \clearfield{number}%
  \clearfield{pages}%
  \clearfield{month}%
  \clearfield{note}%
  \clearfield{issn}%
  \clearfield{doi}%
  \clearlist{publisher}%
  \clearlist{location}%
  \clearlist{editor}%
  \clearname{editor}%
  \clearfield{edition}%
  \clearfield{URL}%
  \clearfield{eprint}%
}

{\end{list}}

\begin{document}
\mainmatter              
\title{The State of Robot Motion Generation\vspace{-.15in}}
\titlerunning{The State of Robot Motion Generation}  

\author{\vspace{-0.05in} Kostas E. Bekris, Joe Doerr, Patrick Meng, Sumanth Tangirala}
\authorrunning{Bekris et al.} 

\institute{Computer Science Dept., Rutgers University, New Brunswick NJ 08901, USA\\
\email{kostas.bekris@cs.rutgers.edu}}

\maketitle              

\vspace{-.2in}
\begin{abstract}
This paper reviews the large spectrum of methods for generating robot motion proposed over the 50 years of robotics research culminating in recent developments. It crosses the boundaries of methodologies, typically not surveyed together, from those that operate over explicit models to those that learn implicit ones. The paper discusses the current state-of-the-art as well as properties of varying methodologies, highlighting opportunities for integration.
\vspace{-.1in}
\keywords{robot motion generation, task and motion planning, control, imitation learning, reinforcement learning,
foundation models}
\end{abstract}

\vspace{-.3in}
\section{Introduction}
\vspace{-.1in}

The robotics community is grappling with a critical question. Will the emerging set of data-driven methods for generating robot motion supersede the traditional techniques as access to robot motion data increases? 

In this context, this paper reviews methods for robot motion generation, which are classified into those that operate given an explicit model vs. those that implicitly learn one from data. Explicit models can correspond to analytical expressions for the world geometry and dynamics or an explainable, numerical approximation in the form of a simulator. Motion generation given explicit models is rather mature and methods are being deployed on real systems, such as autonomous vehicles and industrial manipulators. At the same time, there is increasing excitement for data-driven methods, which have been demonstrated to perform complex tasks, such as dexterous manipulation and unstructured locomotion. These methods often do not depend on explicit models. Instead, they learn implicit representations, which are stored in the internal parameters of machine learning models.

Given the challenge of comprehensively reviewing the vast amount of work in this area across disciplinary boundaries, the focus is on breadth rather than diving deeply into specific methodologies. Similarly, the focus is on principles that are applicable across robotic platforms instead of techniques that are specific to certain hardware configurations. 

The paper concludes with a discussion regarding the properties of the various robot motion generation methods. It argues that integrative approaches and closer interactions between different sub-communities can help develop more robust, safe solutions that can be reliably deployed at reasonable costs and human engineering effort.




\vspace{-.15in}
\section{Motion Generation given Explicit Models}
\vspace{-.1in}

\begin{figure}[h]
\centering
\vspace{-.275in}
\includegraphics[width=0.9\textwidth]{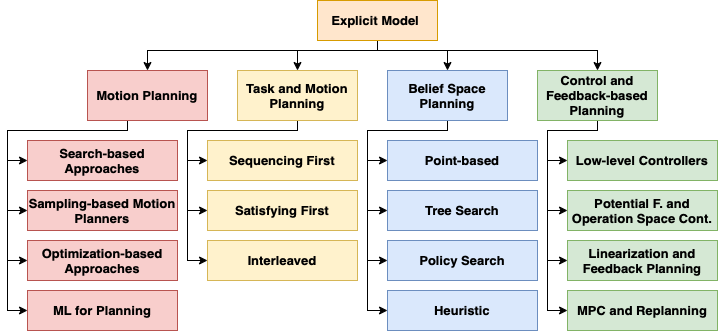}
\vspace{-.175in}
\caption{Robot motion generation methods that operate over an explicit model.}
\label{fig:explicit_models}
\vspace{-.35in}
\end{figure}

Figure \ref{fig:explicit_models} classifies methods that operate over an explicit model. {\bf Motion planning} methods generate safe nominal paths/trajectories to a goal given a fully-observable world model. {\bf Task and motion planning} extends the principle to tasks that require sequencing multiple goals. Such solutions can be executed open-loop if the underlying model is accurate. Robot models, however, are imperfect, resulting in failures upon execution. Given this challenge, {\bf planning under uncertainty} methods aim to compute robust policies to  disturbances that can be modeled. Alternatively, {\bf control and feedback-based planning} methods tightly integrate perception and motion generation so that the robot dynamically reacts to deviations from desired behavior given observations.  


\vspace{-.1in}
\subsection{Motion Planning}
\vspace{-.1in}


Motion planning aims to identify a path with minimal cost, such as the shortest path or the fastest trajectory that brings a robot to a desirable goal state, without undesirable collisions, given a fully-observable world model.



{\bf Search-based Approaches:} Uninformed search methods, such as Uniform Cost Search (UCS) or Dijkstra's algorithm \cite{dijkstra}, can compute optimal paths over a discrete representation of the state space in the form of a grid or a graph given a cost function. Informed alternatives, such as A$^*$ \cite{A*}, utilize heuristic cost-to-go estimates from each state to accelerate solution discovery and can still return the optimum solution over the discrete representation for an admissible/consistent heuristic. Due to the comprehensive nature of search and the discrete representation, search methods suffer from the curse of dimensionality, i.e., the possible states to be explored grow exponentially with dimensions, rendering them computationally infeasible for many robotics problems given na\"ive discretizations. There have been many successful applications of search methods in robotics, however, such as planning for autonomous vehicles \cite{likhachev2009planning} and single or dual arm planning \cite{likhachev2013}.

{\bf Sampling-based Motion Planners (SBMPs):} Sampling provides graph-based representations for searching the collision-free subset of a robot's state space in a more scalable manner than grids. The Probabilistic Roadmap Method (PRM) \cite{PRM} samples collision-free configurations as nodes of a roadmap, and collision-free local paths define the edges. The roadmap can then be used to solve multiple queries via search. For specific queries, the Rapidly Exploring Random Tree (RRT) \cite{RRT} generates a tree data structure rooted in the robot's start state to quickly explore the free state space until it reaches the goal's vicinity. RRT does not require a ``steering function'', i.e., the ability to perfectly connect two robot states, a primitive required by other methods, allowing it to deal with dynamical systems where ``steering functions'' are unavailable. PRM and RRT are provably suboptimal, and asymptotically optimal variants (PRM*, RRT*) \cite{PRM*RRT*} guarantee that the discovered paths converge to optimal ones as sampling progresses. Recent planners \cite{AO-sbmp} \cite{AO-A} also achieve asymptotic optimality for kinodynamic problems. SBMPs have been applied to autonomous driving \cite{self-driving} and manipulation \cite{SB_manipulation}. The Open Motion Planning Library (OMPL) provides software implementations for most SBMPs \cite{OMPL}.

{\bf Optimization-based Approaches:} The above methods are comprehensive and aim to explore the entire feasible state space, which can lead to increased solution times. Moreover, their basic instances don't utilize gradient information. The alternative is to locally optimize paths or trajectories for robotic systems given an objective function under physical or operational limits, such as collision avoidance and control bounds. Covariant Hamiltonian Optimization for Motion Planning (CHOMP) \cite{CHOMP} demonstrated this principle by iteratively optimizing paths by reducing collision and trajectory costs. TrajOpt \cite{TrajOpt} utilizes sequential convex optimization and progressively ensures that each iteration produces feasible and collision-free trajectories. k-Order Markov Optimization (KOMO) \cite{KOMO} treats trajectory optimization as a sparse nonlinear program and tries to address high-dim. problems by leveraging sparsity in dynamics and constraints. Factor graphs, a graphical optimization tool rooted in state estimation, and least-squares optimization can be also used for trajectory optimization \cite{steap}. A recent approach, the Graph of Convex Sets \cite{GCS}, combines optimization and SBMPs. It builds a graphical structure where nodes are convex regions of the free space, and optimization finds a trajectory over the set of nodes that connects the start with the goal. When optimization techniques work, they tend to find high-quality solutions fast. They can suffer, however, from local minima, which arise from the nonlinear and non-convex nature of robotics problems.

{\bf Machine Learning (ML) for Planning:} ML can be used to improve the computational efficiency of planning \cite{surveyMLSBMP}. Some approaches focus on planning components, such as effective sampling \cite{modelCspace}, avoiding collisions \cite{learned_bias}, or distance metrics \cite{distance_metric}. ML can also determine which combination of methods is best suited for a particular problem \cite{planner_selection}, or when a solution is not feasible \cite{meta-reasoning}. \emph{Neural Motion Planning (NMP)} \cite{MPNet} employs neural networks to approximate a planner's operation typically given data from a simulator. An encoder processes environmental data, like point clouds, to create a compact latent space representation. Then, a planning network predicts the robot's next configuration based on its current state, the goal state and the encoded environment. 






\vspace{-.15in}
\subsection{Task and Motion Planning (TAMP)} \label{TAMP}
\vspace{-.1in}


TAMP methods target long-horizon and multi-step robotic tasks, which may include moving through a sequence of goals or manipulating the environment \cite{TAMPSurvey}. TAMP methods typically define low-level operators with motion constraints and high-level logical relationships between the operators. Operators can contain hybrid discrete and continuous parameters, motion constraints, preconditions, and effects, which make use of manually defined lifted variables. Planning is then performed via search across these lifted states. Possible state transitions are operators with satisfied preconditions, resulting in a sequence of operators called a plan skeleton. The hybrid parameters in the plan skeleton (e.g., start and goals) must be solved along with the low level operator trajectories. TAMP techniques can be categorized by the order they sequence and satisfy operators \cite{TAMPSurvey}:  {\bf Sequencing first} \cite{TAMPSequencingFirst} defines a plan skeleton, i.e., a sequence of operators without satisfying their preconditions; this can cause frequent infeasible plans due to the lifted variables not being descriptive enough of underlying constraints. {\bf Satisfying first} trajectories for operators, then planning with them, can solve this problem, but can spend time creating useless satisfied operators due to not having a plan skeleton to direct search \cite{TAMPSatisfactionFirst}. {\bf Interleaved} sequencing and satisfying methods can provide a mix of both by checking low-level information for feasibility during task planning \cite{TAMPInterleaved}. TAMP critically relies on engineering to define preconditions and effects that properly characterize operator behavior and expressive lifted variables. Basic TAMP approaches also struggle under partial observability and uncertainty, which is the focus of recent efforts \cite{TAMPURA}.





\vspace{-.15in}
\subsection{Belief Space Planning}
\vspace{-.1in}

The above methods assume a deterministic, perfect world model. Sensing noise and inaccurate execution introduces uncertainty, however, and the need to generate robust motions to such disturbances. If these noise sources can be modeled probabilistically, (Partially Observable) Markov Decision Processes (PO)MDPs provide a formulation to reason about uncertainty. A Markov Decision Process (MDP) consists of a set of states $S$, set of actions $A$, transition probabilities between states $T(s_{t+1} \mid s_t, a_t)$, and a reward function $R(s_t,a_t,s_{t+1})$. (PO)MDPs employ belief distributions for the problem representation, i.e., probability distribution over states given a set of observations $O$. They can be integrated with Bayesian state estimation (e.g., Kalman or particle filters) that return such beliefs. Solutions to (PO)MDPs are not nominal paths but policies that map beliefs to actions. Due to the consideration of uncertainty, computing an exact solution to a (PO)MDP is often computationally intractable \cite{Kurniawati}, especially given the continuous state and action spaces of robotics. 


This has motivated approximate solutions. The Successive Approximations of the Reachable Space under Optimal Policies (SARSOP) \cite{PBVI} applies {\bf point-based} value iteration and offline sampling to focus on representative beliefs and determine the best action given the sampled set. Determinized Sparse Partially Observable Trees (DESPOT) \cite{DESPOT} employ {\bf tree search} online to compute the optimal action. Online methods can also perform {\bf policy search} by using a controller in a limited search space, increasing scalability but not bounding solution quality \cite{MCVI}. {\bf Heuristic} rules or assumptions can reduce complexity by ignoring long-term consequences and focus on immediate gains or most likely outcomes. For instance, Pre-image Back Chaining constructs state-action sequences (pre-images) leading backward from the goal \cite{pre-image}. Generalized Belief Space (GBS) planning dynamically adapts to ongoing sensing updates \cite{GBS}.

\vspace{-.15in}
\subsection{Control and Feedback-based Planning}
\vspace{-.1in}



Instead of explicitly modeling uncertainty, control tightly integrates  state estimation and motion generation in a closed-loop. Thus, it defines policies that are reactive to different possible outcomes that may be observed upon execution. 

{\bf Low-level Controllers:} \emph{Proportional Integral Derivative (PID)} control and related tools are simple feedback strategies that are robust once tuned. They are ubiquitous for tracking desirable robot controls from higher-level motion generation processes. \emph{Path Tracking Controllers} dynamically select controls given the latest state estimate to minimize path deviation. These controllers, however, are myopic and typically neither reason about the desired goal or obstacles.






{\bf Potential Functions and Operational Space Control:} \emph{Potential functions} \cite{KhatibPotFunc1985} define attractive fields towards the goal and repulsive ones that push away from obstacles. Moving along the negative gradient of the sum of these fields provides the motion vector. Some engineering is needed to tune the parameters of the potentials. In complex environments, the corresponding potential may have multiple minima and goal convergence is not guaranteed. \emph{Navigation functions} \cite{RimonNavigationFunc1988} are smooth and ensure a single minimum at the goal, but they are more complicated to design and can only be constructed for specific environments (i.e., sphere and star worlds). Such control policies do not have to be defined directly in the robot's state space. \emph{Operational Space Control} \cite{Khatib_operational_space} defines such control laws in lower-dim. task spaces. For instance, if the task involves the robot's end-effector, a control law is defined to move it towards a desired goal unencumbered by other robot constraints. Additional control laws can be defined so that links avoid collisions. The corresponding motion vectors for the individual links are then mapped and integrated to a state space motion for the robot via the pseudo-inverses of the robot's Jacobian matrix, which relates robot joint velocities to link velocities through a linear transformation parameterized by the joint states. This allows for \emph{multi-level hierarchical control} \cite{sentis_hierarchical}, where a hierarchy can be imposed over constraints, operational tasks, and soft objectives. Then, lower priority objectives are solved in the null space of higher priority ones once projected to the state space. These strategies rely on precise robot models and high-quality sensing to provide accurate feedback on robot state. 


{\bf Linearization and Feedback-based Planning:} Principles of linear control can be applied for robot motion generation. The \emph{Linear Quadratic Regulator (LQR)} provides an optimal solution for linear time-invariant systems given a quadratic cost function as a control law of the form $u(t)= - K \cdot x(t)$, where $x(t)$ and $u(t)$ are respectively the system's state and the control to be applied at time $t$. Most robots, however, are nonlinear and time-varying, while tasks involve complex, non-quadratic objectives. \emph{Feedback linearization} locally transforms nonlinear systems into equivalent linear ones so that linear control laws can be applied. It can be used for tracking states $x_d$ along a desired trajectory by minimizing the error $e = x(t) - x_d(t)$. Given the linearization, these solutions tend to work in the vicinity of the desired goal/trajectory. They can be ineffective when the feedback-linearized system behaves very differently from the original nonlinear system. To expand the set of initial conditions from which the goal can be reached, \emph{sequential composition} of such feedback policies \cite{FunnelComposition1999} can give rise to hybrid controllers that sequentially switch between them. \emph{LQR-Trees} \cite{lqrTreesTedrake2010} apply sequential composition by combining linear control and SBMPs. They use control verification to evaluate the region of attraction (RoA) of local LQR controllers. They expand a tree backwards from the goal and sample controllers that bring the robot to the RoAs of controllers that lead to the goal. They have been applied on dynamic environments \cite{jaffar2022pip} and for systems with dynamics \cite{verginis2022kdf}.





{\bf Model-Predictive Control (MPC) and Replanning:} MPC is a simple but powerful feedback strategy that aims to solve fast and repetitively finite horizon optimization problems given the latest state observations \cite{camacho2013model}. MPC executes the initial portion of the motion and then updates given the latest observation. During each step, MPC uses the system's model to predict future behavior and makes control decisions that minimize a cost function at the end of the finite horizon, while adhering to constraints, such as collision avoidance. Initial MPC approaches used analytical methods to predict future states, and Linear MPC \cite{linear-mpc} utilizes linear analytical models or approximations. Various Non-Linear MPC (NMPC) variants have been proposed \cite{npmpc}. Shooting MPC is a common NMPC approach that employs numerical methods, such as guessing an initial set of controls and then locally optimizing to minimize the cost function \cite{realtime_mpc}. It is applicable across robotic systems for bridging the model gap. At the same time, shooting MPC involves numerous parameters that require tuning to achieve good performance \cite{MPCTuning2021}. It is also possible to replan using longer horizon planners, where it is important to minimize computational costs so as to achieve tight feedback. For instance, D$^*$ Lite \cite{D*lite}, a replanning variant of A$^*$, updates the path it initially creates to adjust to changes in the workspace. Similarly, replanning versions of SBMPs reuse prior computations to ensure the robot reacts swiftly to changes while maintaining safety \cite{randomizedKDReplan}, including for systems with dynamics \cite{GreedyButSafeKostas2007}.

\vspace{-.15in}
\section{Data-driven Motion Generation with Implicit Models}
\vspace{-.1in}

\begin{figure}[h]
\centering
\includegraphics[width=0.9\textwidth]{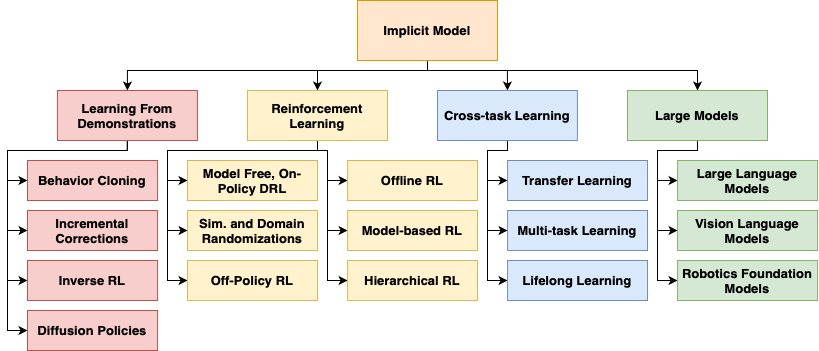}
\vspace{-.15in}
\caption{Robot motion generation methods that operate over an implicit model.}
\label{fig:implicit_models}
\vspace{-.25in}
\end{figure}

Figure \ref{fig:implicit_models} presents a classification of data-driven robot motion generation methods that implicitly learn a model. {\bf Learning from demonstration} methods employ supervision, where a robot is tasked to best replicate the demonstrated behavior. Alternatively, {\bf reinforcement learning} learns to make decisions by experiencing rewards or penalties after interaction with the environment. {\bf Cross-task learning} transfers knowledge from an existing solution to a new task or adapts it dynamically. Finally, {\bf large models}, given access to significant data, allow pre-trained ML models to be fine-tuned and deployed in diverse domains.

\vspace{-.15in}
\subsection{Learning from Demonstrations}
\vspace{-.1in}

Imitation learning mimics the decision-making process given demonstration data.

{\bf Behavior Cloning (BC)} trains a model in a supervised manner given a dataset of demonstrations to provide a motion policy. Successful applications of BC have enabled robots to learn complex behaviors \cite{effectiveBC}\cite{teleoperation}. The process starts with data collection, which includes capturing sensory inputs and corresponding control actions for a target task. An appropriate ML model architecture (e.g., Convolutional Neural Networks (CNNs) for visual data or Recurrent Neural Networks (RNNs) for sequential data) is trained using supervision to minimize a loss function, such as mean squared error. Gaps between the demonstration and the execution setup, as well as noise and stochasticity, can cause a compounding distributional shift between training and testing \cite{dart}. Noise modulates the perception of the true state and can cause erroneous actions from the BC policy. Stochastic state transitions can move the process from an in-distribution state to an out-of-distribution (OOD) one. These effects can compound and make the policy increasingly generate improper actions bringing the robot's state further OOD. Furthermore, BC is sensitive to engineering decisions, such as observation space, hyperparameters, and recurrent information \cite{StanfordImitationPaper}. These shortcomings motivate more sophisticated approaches.

{\bf Incremental Corrections:} The DAgger algorithm \cite{DAgger} focuses on decision making in OOD states given the available demonstrations. A dataset of rollouts is collected by executing the BC policy. Then, an expert, such as a human or a planner, is queried to provide the proper actions at the states encountered in this dataset, and the policy is retrained. This process is iterative, as new OOD states will be encountered after each retraining. DAgger heavily relies on an expert to provide the corrections. SafeDAgger provided a more query-efficient extension \cite{safedagger}. It first uses a safety policy to predict the error of the learned policy. This safety policy selects only a small subset of training examples to be collected. 





{\bf Inverse Reinforcement Learning (IRL)} focuses on extracting the problem's underlying dense reward function given demonstrations to achieve improved generalization relative to BC \cite{IRLSurvey}. This reward function can then be used to extract the policy provided an MDP problem formulation. Initial works \cite{Ng2000AlgIRL} modeled the reward as a linear combination of input features, and the weights of this linear combination are obtained by solving an optimization problem. A significant challenge of IRL, however, is the difficulty in scaling to high-dim. systems, which recent approaches attempt to mitigate \cite{RecedingHorizonIRL}.

{\bf Diffusion Policies:} Demonstrations commonly include multi-modal behaviors, which multi-layer perceptrons (MLP) may find difficult to express since they are trained via mean squared error (MSE) that averages the demonstrated actions resulting in erroneous choices \cite{diffusionImitation}. In compuater vision, diffusion models have become popular for generating images by expressing multi-modal distributions \cite{ddpm}. This motivates diffusion processes as implicit models for BC, where Diffusion Policy \cite{diffusionpolicy} has outperformed alternatives, such as LSTM-GMM \cite{lstm-gmm} in learning from demonstrations. Inference speed of diffusion policies is a challenge for robot motion generations \cite{efficientDiffPolicy} and is an active area of research.


\vspace{-.15in}
\subsection{Reinforcement Learning (RL)}
\vspace{-.1in}


One area of robotics that has garnered interest \cite{safe_learning} is Reinforcement Learning (RL), where a machine learning model predicts the Q function of an MDP and the corresponding policy. It has been applied across tasks, e.g., grasping \cite{QtOpt}, locomotion \cite{deeploco}, and assembly \cite{successfulAssembly}. 


{\bf Model-free, On-Policy RL:} Standard RL gathers data from interaction with the environment, where Policy Gradient (PG) methods generate a batch of full trajectory data, compute the gradient of each state transition, and scale each gradient by the respective discounted return. PGs update the policy and repeat the process towards maximizing the expected cumulative reward. Proximal Policy Optimization (PPO) \cite{PPO} is a PG method that leverages a value function, which is the average discounted return from a given state. These are on-policy methods, i.e., they only use data collected from the most recent policy. Exploration can be achieved through the periodic execution of random actions (epsilon greedy), injecting Gaussian noise to policy outputs \cite{DDPG}, or using a stochastic policy for entropy maximization \cite{PPO}. 


RL faces \emph{multiple challenges} in robotics: (i) \underline{sample inefficiency}: Policies need a lot of data to train and take significant training time. Real robot data, however, can be slow, expensive, and unsafe to collect \cite{lessonslearnedtrainrobot}. (ii) \underline{instability:} Policies can be inconsistent across training sessions due to poorly designed rewards, exploration strategies, learning rate parameters or learning error from neural networks. (iii) \underline {reward engineering:} Simple rewards are sparse, i.e., they assign a 0 reward everywhere except at the goal. While desirable, sparse rewards often do not find a successful behavior in reasonable time \cite{lessonslearnedtrainrobot}. This motivates dense rewards to guide exploration, which require manual engineering. While dense rewards can improve training time and stability, they can also lead to wrong behaviors, if they do not match the true task objective. (iv) \underline {long horizon tasks:} The state space to be explored for such tasks is even larger making the propagation of rewards across subproblems difficult. 

The above challenges have motivated multiple RL variations in robotics:

{\bf Simulation and Domain Randomization:} Simulators, which are explicit world models, are useful for addressing sample inefficiency by generating data without real-world experiments. They suffer, however, from the model gap issue. Even so, it may still be possible to create good policies via \emph{domain randomization (DR)} \cite{domainrandomization}, which can partially compensate for imperfect models during sim-to-real transfer. DR results in behaviors that have the highest expected reward across a variety of underlying dynamics and perception errors. 



{\bf Off-policy RL} \label{offpolicyRL} Sample inefficiency  motivates data reuse. Off-policy RL stores the expected discounted future reward of any in-distribution state-action tuple on a Q function, which is used to train the policy $\pi$. To update the parameters $\theta$ of the Q function for a given state transition $(s,a,r,s'$) (state, action, reward, next state), the immediate reward $r$ and future rewards $Q(s', \pi_{\phi}(s'))$ are used to define the loss function: $L(\theta) = E(Q_{\theta}(s,a) - r + \gamma \cdot Q_{\theta}( s', \pi_{\phi}(s')))^2$. This allows off-policy RL algorithms, such as TD3 \cite{TD3} and SAC \cite{SAC}, to be  more sample efficient than online RL. Off-policy RL, however, is complicated by the instability of neural network learning, especially because errors compound while learning Q functions \cite{DQN}. A way to mitigate instability is to use target Q functions to query for future rewards \cite{TD3}\cite{SAC}\cite{DQN}. The target Q functions average their weights slowly with the learned Q function's weights. Given that the policy is defined as $\pi^*(s) = \arg\max_{a} Q(s, a)$, errors in Q function learning, which overvalue actions, are propagated rapidly. This is called an overestimation bias \cite{TD3}. One way to mitigate overestimation bias to is use a clipped, double Q function, an ensemble of two Q networks coupled with target Q networks, that take the lower value of the two when queried for future rewards to underestimate Q values when uncertain \cite{TD3}\cite{SAC}. Hindsight Experience Replay (HER) \cite{HER} aims for sample-efficiency in multi-goal problems given only sparse rewards. It relabels the end or intermediate states of executed trajectories as the desired goal allowing to train over all available experiences. 





{\bf Offline RL} Offline RL is a subcase of off-policy RL, which doesn't interact online with the environment to modulate the replay buffer. Instead it uses offline data, focusing on sample efficiency. Offline RL aims to improve the optimality of suboptimal demonstrations by patching together data towards the most optimal solution rather than imitating the data \cite{whenuseOfflineRL}. Additionally, it can handle multi-modal demonstrations as it can solve for the single most optimal action for each state. The policy $\pi$ used for updating the Q function is parameterized via a function approximator. This can cause $\pi$ to choose actions that are out-of-distribution (OOD). The Q function queried with the OOD state-action pair will output an untrained value, which can be an erroneous large reward that can propagate and degrade performance \cite{sergeyofflineRLsurvey}. In online RL, when an OOD state-action pair is overvalued, $\pi$ will favor the action and gather data on the OOD state-action pair, which will remedy the error. IQL \cite{IQL}, which is an offline RL method, mitigates distributional shift by using only state-action pairs from the dataset. Instead, CQL \cite{CQL} employs a conservative estimate on Q values.  


{\bf Model-based RL} learns a model of the environment, which is then used to roll out the policy to generate training data offline. Relative to a simulator, the learned model can operate over a more compact state representation, run faster, and can be queried for any given state. The major drawback is that RL may exploit errors in the learned model. MOPO \cite{mopo} aims to mitigate these effects by detecting when the query to the model is out of distribution. 



{\bf Hierarchical RL (HRL)} focuses on long horizon tasks, similar to TAMP (Section \ref{TAMP}). It temporally abstracts high-level actions that correspond to low-level learned policy. The high-level actions can target exploration to better cover the space and more easily assign rewards \cite{HRLSurvey}. Using random actions to search is not effective in long horizon tasks and guidance for their generation is critical. Long horizon tasks remain rather challenging for RL approaches.





\vspace{-.15in}
\subsection{Cross-task Learning} 
\vspace{-.15in}


Demonstrations from related or earlier tasks can help bootstrap or guide RL.




{\bf Transfer Learning} methods exhibit high variability. One approach bootstraps RL by using an adaptively-weighted auxiliary term in the loss function of PPO to increase action similarity for the learned policy against the demonstrated one from a task with similar MDP \cite{kickstartingRL}. It is also possible to pre-train multiple tasks with offline RL given a single task-conditioned policy \cite{Dontstartfromscratch}. Online, the approach fine-tunes both the conditioning parameter and policy to automatically reset tasks for autonomous real-world training. For transferring information across heterogeneous MDPs, researchers have mapped states and actions across MDPs \cite{InvariantFeatureSpace} and transfered useful representations across domains \cite{pushingthelimitsofcrossembodiment}. 





{\bf Multi-Task Learning} trains many tasks in parallel while sharing information across tasks to accelerate training and improve generalization \cite{MultitaskLearningSurvey}. Methods focus on how to select parameters to sharing so as to effectively transfer learned representations between tasks \cite{multitaskadashare}. 


{\bf Lifelong Learning} emphasizes learning a new task in a sequence by transferring information from previously learned ones without forgetting how to solve them. Retaining previous task information can be done by mixing previous data with new ones during retraining \cite{lifelonglearningexperiencereplay}. Functionally composed modular lightweight networks have been proposed to learn a large variety of combinatorially related tasks to solve novel combination tasks in a zero-shot manner \cite{mendezmendez}.





\vspace{-.25in}
\subsection{Large Models}
\vspace{-.15in}


{\bf Large Language Models (LLMs)} pretrained on internet-scale data have opened new avenues for robot motion generation. LLMs are capable of semantic reasoning and planning, making them candidates for extracting task specifications and creating action models for task planning \cite{TAMPInterpret}. LLMs have also been used to improve existing action models to handle failure cases \cite{LASP}. These methods often do not generate constraints for low level motion in the operators as in TAMP. Furthermore, they have been used in conjunction with evolutionary optimization as code generators for defining rewards, which can be used to acquire complex skills via reinforcement learning \cite{ma2023eureka}.



{\bf Vision Language Models} (VLMs) integrate information from visual input and language. Saycan \cite{LLMHighlevelreasoner} integrates a visual affordance model, which evaluates possible robot actions, with a language model that interprets the user's commands and generates high-level action plans. VLMs have also been used for autonomous generation of demonstration data over diverse tasks \cite{VLMManipulate}. 


This lead to vision-language-action models (VLAs) or {\bf Robotics Foundation Models}, i.e., large pretrained models that map actions to sensing data and language specifications. They can be fine-tuned for specific tasks to provide improvements in training time and generalization over trainingfrom scratch. OpenVLA \cite{OpenVLA} is such a model trained on a large robotic manipulation dataset \cite{hugerobottrajectorydataset} and utilizes large pretrained vision encoders and language models. It has been argued that it performs well on in-distribution tasks and robotic embodiments, while it can be fine-tuned for novel tasks and robotic embodiments.




\vspace{-.15in}
\section{Perspectives on Robot Motion Generation Research}
\vspace{-.1in}

{\bf Promises and Pitfalls of the Explicit Model Approach:} These methods have a long history and can reliably solve various challenges today. For instance, planning collision-free motions for industrial arms is reliably addressed today at high speeds. Integration with state estimation allows mobile robots to execute navigation in semi-structured domains reliably. Challenges arise where model or state estimation reliability is low. Examples of such setups often involve the presence of complex contacts and partial observability, such as the manipulation of previously unknown objects in clutter, locomotion over uneven terrains, navigation at high speeds or in dynamic, unstructured environments. 

For long-horizon tasks, TAMP methods require an engineer to encode possible pathways and concepts before symbolic reasoning, which often involves expensive combinatorial reasoning. These methods are limited to the variables defined and cannot easily anticipate changes from what has been programmed. While belief-space planning targets noise and partial observability, it requires significant computation and access to accurate models of uncertainty, which are not always available. At the same time, feedback-based solutions do find applicability in real-world domains, either via MPC or hierarchical control strategies but fully understanding the conditions under which a controller can solve a problem is an active challenge. 

{\bf Promises and Pitfalls of the Implicit Model Approach:} The progress in ML promises effective data-driven motion generation methods that can access prior experience and do not require an engineered model or even accurate state estimation. They have unblocked perception tasks, such as object detection and estimation, which are often prerequisites for robust motion generation. Learning from demonstration, especially with diffusion processes that allow reflecting multi-modal distributions, can achieve impressive results exactly in tasks that the traditional, explicit model approaches face challenges with, such as dexterous manipulation and locomotion. This is true, however, as long as the setup upon execution resembles the demonstration setup, thus limiting generalization. The various versions of reinforcement learning bring the promise of broader generalization, and there have been many successful demonstrations of learning skills for robotic tasks via RL, though sample inefficiency still remains a bottleneck for achieving highly accurate solutions across a wide set of initial conditions. 

These limitations have motivated the robotics community to pursue the direction of collecting more data for robotics problems in lab environments, which may be diverse across embodiments and tasks \cite{hugerobottrajectorydataset}, towards the objective of mimicking the success of foundation models in language and vision challenges. While this direction should be pursued, it is not clear that it is possible to collect internet-scale demonstration data that will allow learning robust enough policies that cover the space of possible tasks that robots need to solve in novel, unstructured, and human environments. Furthermore, predicting when the resulting solutions will be successful is challenging, which is a significant concern, as failures in robotics can cause physical harm. 

{\bf Integrative Directions:} There is promise in the integration of solutions. For instance, data-driven methods training can benefit from simulation, where the explicit model approaches act as the demonstrators under full observability. The data-driven methods can learn to map sensing data to the actions demonstrated by the planners. This requires accurate enough models for the policies to be transferable to real systems, as well as strategies that makes these policies robust to varying conditions. 

Upon deployment, data-driven methods must be part of architectures that provide safety and verification, where they can benefit from explicit model methods. For instance, describing when a learned controller is successful, similar to control verification, can allow safe deployment. It can also assist in controller composition for long-horizon tasks, where high-level symbolic reasoning can be beneficial. Such task planning needs to be adaptive and allow a robot to dynamically define pre/postconditions, and discover new skills for its task. It should also be accompanied by failure explanation to identify why a problem is not solvable and guide data collection or reasoning for addressing similar challenges in the future. For explainability, maintaining an internal world model or a simulation (e.g., a ``cognitive, physical engine'') that is learned from data and first principles can be useful. There are already successful instances of integrating explicit and implicit models. There is work on natural-langauge-task-specification TAMP problems on large scenes where task plans from an LLM are verified using a model and pose-level planning is performed by a classical planner \cite{sayplan}. Another integration approach queries a VLM to solve a TAMP problem on the fly by creating 3D key points that resultin in a task plan as a sequence of path and goal constraints that guide motion planning \cite{rekep}.

A gap towards integrative solutions is the lack of common interfaces, software components and benchmarks that would allow to easily switch and experiment with components from different methodologies. Most existing instances of software infrastructure either support one set of methods or the other, requiring the ad hoc tool composition for a novel integrative approach.

\vspace{-.15in}
\AtNextBibliography{\footnotesize}
\printbibliography

@inproceedings{teleoperation,
  title={Deep imitation learning for complex manipulation tasks from virtual reality teleoperation},
  author={Zhang, Tianhao and McCarthy, Zoe and Jow, Owen and Lee, Dennis and Chen, Xi and Goldberg, Ken and Abbeel, Pieter},
  booktitle={ICRA},
  pages={},
  year={2018}
}

@article{InvariantFeatureSpace,
  title={Learning invariant feature spaces to transfer skills with reinforcement learning},
  author={Gupta, Abhishek and Devin, Coline and Liu, YuXuan and Abbeel, Pieter and Levine, Sergey},
  journal={ICLR},
  year={2017}
}

@inproceedings{Dontstartfromscratch,
  title={Don’t start from scratch: Leveraging prior data to automate robotic reinforcement learning},
  author={Walke, Homer Rich and Yang, Jonathan Heewon and Yu, Albert and Kumar, Aviral and Orbik, J{\k{e}}drzej and Singh, Avi and Levine, Sergey},
  booktitle={CoRL},
  pages={},
  year={2023}
}

@article{kickstartingRL,
  title={Kickstarting deep reinforcement learning},
  author={Schmitt, Simon and Hudson, Jonathan J and Zidek, Augustin and Osindero, Simon and Doersch, Carl and Czarnecki, Wojciech M and Leibo, Joel Z and Kuttler, Heinrich and Zisserman, Andrew and Simonyan, Karen and others},
  journal={ arXiv:1803.03835},
  year={2018}
}

@article{whenuseOfflineRL,
  title={When should we prefer offline reinforcement learning over behavioral cloning?},
  author={Kumar, Aviral and Hong, Joey and Singh, Anikait and Levine, Sergey},
  journal={ICLR},
  year={2022}
}

@article{StanfordImitationPaper,
  title={What matters in learning from offline human demonstrations for robot manipulation},
  author={Mandlekar, Ajay and Xu, Danfei and Wong, Josiah and Nasiriany, Soroush and Wang, Chen and Kulkarni, Rohun and Fei-Fei, Li and Savarese, Silvio and Zhu, Yuke and Mart{\'\i}n-Mart{\'\i}n, Roberto},
  journal={CoRL},
  year={2021}
}

@article{diffusionImitation,
  title={Imitating human behaviour with diffusion models},
  author={Pearce, Tim and Rashid, Tabish and Kanervisto, Anssi and Bignell, Dave and Sun, Mingfei and Georgescu, Raluca and Macua, Sergio Valcarcel and Tan, Shan Zheng and Momennejad, Ida and Hofmann, Katja and others},
  journal={ICLR},
  year={2023}
}

@inproceedings{dart,
  title={Dart: Noise injection for robust imitation learning},
  author={Laskey, Michael and Lee, Jonathan and Fox, Roy and Dragan, Anca and Goldberg, Ken},
  booktitle={CoRL},
  pages={},
  year={2017}
}

@article{QtOpt,
  title={Learning hand-eye coordination for robotic grasping with deep learning and large-scale data collection},
  author={Levine, Sergey and Pastor, Peter and Krizhevsky, Alex and Ibarz, Julian and Quillen, Deirdre},
  journal={IJRR},
  volume={},
  number={},
  pages={},
  year={2018},
  publisher={SAGE Publications Sage UK: London, England}
}

@article{deeploco,
  title={Deeploco: Dynamic locomotion skills using hierarchical deep reinforcement learning},
  author={Peng, Xue Bin and Berseth, Glen and Yin, KangKang and Van De Panne, Michiel},
  journal={TOG},
  volume={},
  number={},
  pages={},
  year={2017}
}

@article{mopo,
  title={Mopo: Model-based offline policy optimization},
  author={Yu, Tianhe and Thomas, Garrett and Yu, Lantao and Ermon, Stefano and Zou, James Y and Levine, Sergey and Finn, Chelsea and Ma, Tengyu},
  journal={NeurIPS},
  volume={},
  pages={},
  year={2020}
}

@inproceedings{DAgger,
  title={A reduction of imitation learning and structured prediction to no-regret online learning},
  author={Ross, St{\'e}phane and Gordon, Geoffrey and Bagnell, Drew},
  booktitle={AISTATS},
  pages={},
  year={2011}
}

@article{TAMPInterpret,
  title={InterPreT: Interactive Predicate Learning from Language Feedback for Task Planning},
  author={Han, Muzhi and Zhu, Yifeng and Zhu, Song-Chun and Wu, Ying Nian and Zhu, Yuke},
  journal={RSS},
  year={2024}
}

@article{LASP,
  title={Language-Augmented Symbolic Planner for Open-World Task Planning},
  author={Chen, Guanqi and Yang, Lei and Jia, Ruixing and Hu, Zhe and Chen, Yizhou and Zhang, Wei and Wang, Wenping and Pan, Jia},
  journal={RSS},
  year={2024}
}

@article{hugerobottrajectorydataset,
  title={{Open X-embodiment: Robotic learning datasets and RT-X models}},
  author={Padalkar, Abhishek and Pooley, Acorn and Jain, Ajinkya and Bewley, Alex and Herzog, Alex and Irpan, Alex and Khazatsky, Alexander and Rai, Anant and Singh, Anikait and Brohan, Anthony and others},
  journal={arXiv preprint arXiv:2310.08864},
  year={2023}
}

@article{OpenVLA,
  title={{OpenVLA: An Open-Source Vision-Language-Action Model}},
  author={Kim, Moo Jin and Pertsch, Karl and Karamcheti, Siddharth and Xiao, Ted and Balakrishna, Ashwin and Nair, Suraj and Rafailov, Rafael and Foster, Ethan and Lam, Grace and Sanketi, Pannag and others},
  journal={arXiv:2406.09246},
  year={2024}
}

@article{TAMPURA,
  title={Partially Observable Task and Motion Planning with Uncertainty and Risk Awareness},
  author={Curtis, Aidan and Matheos, George and Gothoskar, Nishad and Mansinghka, Vikash and Tenenbaum, Joshua and Lozano-P{\'e}rez, Tom{\'a}s and Kaelbling, Leslie Pack},
  journal={RSS},
  year={2024}
}

@article{IQL,
  title={Offline reinforcement learning with implicit q-learning},
  author={Kostrikov, Ilya and Nair, Ashvin and Levine, Sergey},
  journal={ arXiv:2110.06169},
  year={2021}
}

@article{CQL,
  title={Conservative q-learning for offline reinforcement learning},
  author={Kumar, Aviral and Zhou, Aurick and Tucker, George and Levine, Sergey},
  journal={NeurIPS},
  volume={},
  pages={},
  year={2020}
}

@article{HRLSurvey,
  title={Hierarchical reinforcement learning: A comprehensive survey},
  author={Pateria, Shubham and Subagdja, Budhitama and Tan, Ah-hwee and Quek, Chai},
  journal={CSUR},
  volume={},
  number={},
  pages={},
  year={2021}
}

@inproceedings{successfulAssembly,
  title={Deep reinforcement learning for high precision assembly tasks},
  author={Inoue, Tadanobu and De Magistris, Giovanni and Munawar, Asim and Yokoya, Tsuyoshi and Tachibana, Ryuki},
  booktitle={IROS},
  pages={},
  year={2017}
}

@article{DQN,
  title={Human-level control through deep reinforcement learning},
  author={Volodymyr Mnih and Koray Kavukcuoglu and David Silver and Andrei A. Rusu and Joel Veness and Marc G. Bellemare and Alex Graves and Martin A. Riedmiller and Andreas Kirkeby Fidjeland and Georg Ostrovski and Stig Petersen and Charlie Beattie and Amir Sadik and Ioannis Antonoglou and Helen King and Dharshan Kumaran and Daan Wierstra and Shane Legg and Demis Hassabis},
  journal={Nature},
  volume={},
  number={},
  pages={},
  year={2015},
  publisher={Nature Publishing Group UK London}
}

@inproceedings{TD3,
  title={Addressing function approximation error in actor-critic methods},
  author={Fujimoto, Scott and Hoof, Herke and Meger, David},
  booktitle={ICML},
  pages={},
  year={2018}
}

@inproceedings{SAC,
  title={Soft actor-critic: Off-policy maximum entropy deep reinforcement learning with a stochastic actor},
  author={Haarnoja, Tuomas and Zhou, Aurick and Abbeel, Pieter and Levine, Sergey},
  booktitle={ICML},
  pages={},
  year={2018}
}

@article{DDPG,
  title={Continuous control with deep reinforcement learning},
  author={Timothy P. Lillicrap and Jonathan J. Hunt and Alexander Pritzel and Nicolas Heess and Tom Erez and Yuval Tassa and David Silver and Daan Wierstra},
  journal={CoRR},
  year={2015}
}

@article{mendezmendez,
  title={Modular lifelong reinforcement learning via neural composition},
  author={Mendez, Jorge A and van Seijen, Harm and Eaton, Eric},
  journal={ arXiv:2207.00429},
  year={2022}
}

@article{multitaskadashare,
  title={Adashare: Learning what to share for efficient deep multi-task learning},
  author={Sun, Ximeng and Panda, Rameswar and Feris, Rogerio and Saenko, Kate},
  journal={NeurIPS},
  volume={},
  pages={},
  year={2020}
}

@article{MultitaskLearningSurvey,
  title={A survey on multi-task learning},
  author={Zhang, Yu and Yang, Qiang},
  journal={TKDE},
  volume={},
  number={},
  pages={},
  year={2021}
}

@article{lifelonglearningexperiencereplay,
  title={Experience replay for continual learning},
  author={David Rolnick and Arun Ahuja and Jonathan Schwarz and Timothy P. Lillicrap and Greg Wayne},
  journal={NeurIPS},
  volume={},
  year={2019}
}

@article{sergeyofflineRLsurvey,
  title={Offline reinforcement learning: Tutorial, review, and perspectives on open problems},
  author={Levine, Sergey and Kumar, Aviral and Tucker, George and Fu, Justin},
  journal={arXiv:2005.01643},
  year={2020}
}

@article{HER,
  title={Hindsight experience replay},
  author={Marcin Andrychowicz and Filip Wolski and Alex Ray and Jonas Schneider and Rachel Fong and Peter Welinder and Bob McGrew and Josh Tobin and Pieter Abbeel and Wojciech Zaremba},
  journal={NeurIPS},
  volume={},
  year={2017}
}

@article{lessonslearnedtrainrobot,
  title={How to train your robot with deep reinforcement learning: lessons we have learned},
  author={Ibarz, Julian and Tan, Jie and Finn, Chelsea and Kalakrishnan, Mrinal and Pastor, Peter and Levine, Sergey},
  journal={IJRR},
  volume={},
  number={},
  pages={},
  year={2021},
  publisher={SAGE Publications Sage UK: London, England}
}

@article{PPO,
  title={Proximal policy optimization algorithms},
  author={John Schulman and Filip Wolski and Prafulla Dhariwal and Alec Radford and Oleg Klimov},
  journal={ arXiv:1707.06347},
  year={2017}
}

@article{pushingthelimitsofcrossembodiment,
  title={Pushing the limits of cross-embodiment learning for manipulation and navigation},
  author={Jonathan Yang and Catherine Glossop and Arjun Bhorkar and Dhruv Shah and Quan Vuong and Chelsea Finn and Dorsa Sadigh and Sergey Levine},
  journal={RSS},
  year={2024}
}

@article{TAMPSurvey,
  title={Integrated task and motion planning},
  author={Garrett, Caelan Reed and Chitnis, Rohan and Holladay, Rachel and Kim, Beomjoon and Silver, Tom and Kaelbling, Leslie Pack and Lozano-Pérez, Tomás},
  journal={{Annual Review of Control, Robotics, \& Autonomous Systems}},
  volume={},
  number={},
  pages={},
  year={2021},
  publisher={Annual Reviews}
}

@article{IRLSurvey,
title = {A survey of inverse reinforcement learning: Challenges, methods and progress},
journal = {AIJ},
year = {2021},
volume = {},
pages = {},
year = {},
issn = {},
doi = {},
url = {},
author = {Saurabh Arora and Prashant Doshi}
}

@inproceedings{Ng2000AlgIRL,
author = {Ng, Andrew Y. and Russell, Stuart J.},
title = {Algorithms for Inverse Reinforcement Learning},
year = {2000},
isbn = {},
address = {},
booktitle = {ICML},
pages = {},
numpages = {},
series = {}
}

@inproceedings{RecedingHorizonIRL,
 author = {Xu, Yiqing and Gao, Wei and Hsu, David},
 booktitle = {NeurIPS},
 title = {Receding Horizon Inverse Reinforcement Learning},
 year = {2022}
}

@inproceedings{D*lite,
  title={Improved fast replanning for robot navigation in unknown terrain},
  author={Koenig, Sven and Likhachev, Maxim},
  booktitle={ICRA},
  year={2002}
}

@conference{sentis_hierarchical,
author = {Sentis, Luis and Khatib, Oussama},
title = {Synthesis of whole-body behaviors through hierarchical control of behavioral primitives},
booktitle = {IJHR},
year = 2005
}

@conference{safedagger,
author = {Zhang, Jiakai and Cho, Kyunghyun},
title = {Query-efficient imitation learning for end-to-end simulated driving},
year = {2017},
booktitle = {AAAI},
}

@ARTICLE{A*,
  author={Hart, Peter E. and Nilsson, Nils J. and Raphael, Bertram},
  journal={IEEE TSSC}, 
  title={A Formal Basis for the Heuristic Determination of Minimum Cost Paths}, 
  year={1968},
  volume={4},
  number={2},
  pages={100-107},
  doi={10.1109/TSSC.1968.300136}}

@incollection{dijkstra,
  title={A note on two problems in connexion with graphs},
  author={Dijkstra, Edsger W},
  booktitle={Numerische Mathematik},
  year={1959}
}

@article{PRM,
  title={Probabilistic roadmaps for path planning in high-dimensional configuration spaces},
  author={Kavraki, L.E. and Svestka, P. and Latombe, J.-C. and Overmars, M.H.},
  journal={TRO},
  year = {1996}
}

@conference{RRT,
  author={LaValle, S.M. and Kuffner, J.J.},
  booktitle={ICRA}, 
  title={Randomized kinodynamic planning}, 
  year={1999},
}

@article{AO-A,
  title={Asymptotically optimal planning by feasible kinodynamic planning in a state--cost space},
  author={Hauser, Kris and Zhou, Yilun},
  journal={TRO},
  year={2016}
}

@article{PRM*RRT*,
  title={Sampling-based algorithms for optimal motion planning},
  author={Karaman, Sertac and Frazzoli, Emilio},
  journal={IJRR},
  volume={30},
  number={7},
  pages={846--894},
  year={2011},
}

@article{CHOMP,
author = {Zucker, Matt and Ratliff, Nathan and Dragan, Anca and Pivtoraiko, Mihail and Klingensmith, Matthew and Dellin, Christopher and Bagnell, J. and Srinivasa, Siddhartha},
year = {2013},
month = {08},
pages = {1164-1193},
title = {{CHOMP}: Covariant Hamiltonian optimization for motion planning},
volume = {32},
journal = {IJRR},
}

@conference{TrajOpt,
  title={Finding Locally Optimal, Collision-Free Trajectories with Sequential Convex Optimization},
  author={John Schulman and Jonathan Ho and Alex X. Lee and Ibrahim Awwal and Henry Bradlow and P. Abbeel},
  booktitle={RSS},
  year={2013},
}

@techreport{KOMO,
    author = {Marc Toussaint},
    title = {{Newton methods for k-order Markov Constrained Motion Problems}},
    institution = {arXiv:1407.0414},
    year = {2014}
}

@article{ddpm,
  title={Denoising diffusion probabilistic models},
  author={Ho, Jonathan and Jain, Ajay and Abbeel, Pieter},
  journal={NeurIPS},
  year={2020}
}

@inproceedings{diffusionpolicy,
	title={Diffusion Policy: Visuomotor Policy Learning via Action Diffusion},
	author={Cheng Chi and Zhenjia Xu and Siyuan Feng and Eric Cousineau and Yilun Du and Benjamin Burchfiel and Russ Tedrake and Shuran Song},
	booktitle={RSS},
	year={2023}
}

@INPROCEEDINGS{KhatibPotFunc1985,
  author={Khatib, O.},
  booktitle={ICRA}, 
  title={Real-time obstacle avoidance for manipulators and mobile robots}, 
  year={1985},
}

@INPROCEEDINGS{RimonNavigationFunc1988,
  author={Rimon, E. and Koditschek, D.E.},
  booktitle={TRO}, 
  title={{Exact Robot Navigation Using Artificial Potential Functions}}, 
  year={1992},
}

@inproceedings{MPNet,
  title={Motion planning networks},
  author={Ahmed H. Qureshi and Anthony Simeonov and Mayur J. Bency and Michael C. Yip},
  booktitle={ICRA},
  year={2019},
}

@article{surveyMLSBMP,
   title={A Survey on the Integration of Machine Learning with Sampling-based Motion Planning},
   volume={9},
   journal={FTR},
   publisher={Now Publishers},
   author={McMahon, Troy and Sivaramakrishnan, Aravind and Granados, Edgar and Bekris, Kostas E.},
   year={2022},
   pages={266–327} 
}

@conference{modelCspace,
  author={Arslan, Oktay and Tsiotras, Panagiotis},
  booktitle={IROS}, 
  title={Machine learning guided exploration for sampling-based motion planning algorithms}, 
  year={2015},
  }

@INPROCEEDINGS{learned_bias,
  author={Burns, B. and Brock, O.},
  booktitle={ICRA}, 
  title={Sampling-Based Motion Planning Using Predictive Models}, 
  year={2005},
  volume={},
  number={},
  pages={3120-3125}}

@InProceedings{distance_metric,
author="Chiang, Hao-Tien Lewis
and Faust, Aleksandra
and Sugaya, Satomi
and Tapia, Lydia",
title="Fast Swept Volume Estimation with Deep Learning",
booktitle="WAFR",
year="2020",
address="Cham",
pages="52--68"
}

@Inbook{planner_selection,
author="Morales, Marco
and Tapia, Lydia
and Pearce, Roger
and Rodriguez, Samuel
and Amato, Nancy M.",
title="A Machine Learning Approach for Feature-Sensitive Motion Planning",
bookTitle="WAFR",
year="2005"
}

@inproceedings{meta-reasoning,
author = {Li, Sihui and Dantam, Neil},
year = {2021},
booktitle = {RSS},
title = {Learning Proofs of Motion Planning Infeasibility}}

@article{FunnelComposition1999,
author = {R. R. Burridge and A. A. Rizzi and D. E. Koditschek},
title ={Sequential Composition of Dynamically Dexterous Robot Behaviors},
journal = {IJRR},
year = {1999}}

@article{lqrTreesTedrake2010,
author = {Russ Tedrake and Ian R. Manchester and Mark Tobenkin and John W. Roberts},
title ={LQR-trees: Feedback Motion Planning via Sums-of-Squares Verification},
journal = {IJRR},
year = {2010}}

@article{verginis2022kdf,
  title={Kdf: Kinodynamic motion planning via geometric sampling-based algorithms and funnel control},
  author={Verginis, Christos K and Dimarogonas, Dimos V and Kavraki, Lydia E},
  journal={TRO},
  year={2022}
}

@inproceedings{jaffar2022pip,
  title={Pip-x: Funnel-based online feedback motion planning/replanning in dynamic environments},
  author={Jaffar, Mohamed Khalid M and Otte, Michael},
  booktitle={WAFR},
  year={2022}
}

@INPROCEEDINGS{GreedyButSafeKostas2007,
  author={Bekris, Kostas E. and Kavraki, Lydia E.},
  booktitle={ICRA}, 
  title={Greedy but Safe Replanning under Kinodynamic Constraints}, 
  year={2007}}

@article{randomizedKDReplan,
author = {David Hsu and Robert Kindel and Jean-Claude Latombe and Stephen Rock},
title ={Randomized Kinodynamic Motion Planning with Moving Obstacles},

journal = {IJRR},
year = {2002}}

@INPROCEEDINGS{MPCTuning2021,
  author={Edwards, William and Tang, Gao and Mamakoukas, Giorgos and Murphey, Todd and Hauser, Kris},
  booktitle={ICRA}, 
  title={Automatic Tuning for Data-driven Model Predictive Control}, 
  year={2021}}

@inproceedings{PBVI,
author = {Pineau, Joelle and Gordon, Geoff and Thrun, Sebastian},
title = {Point-based value iteration: an anytime algorithm for POMDPs},
year = {2003},
publisher = {Morgan Kaufmann Publishers Inc.},
booktitle = {IJCAI}
}

@inproceedings{MCVI,
author = {Bai, Haoyu and Hsu, David and Lee, Wee and Vien, Ngo},
year = {2010},
title = {Monte Carlo value iteration for continuous-state POMDPs},
booktitle = {WAFR}
}

@article{Khatib_operational_space,
author={Khatib, Oussama},
title = {{A unified approach for motion and force control of robot manipulators: The operational space formulation}},
journal = "JRA",
year = 1987
}

@article{DESPOT,
   title={DESPOT: Online POMDP Planning with Regularization},
   journal={JAIR},
   publisher={AI Access Foundation},
   author={Ye, Nan and Somani, Adhiraj and Hsu, David and Lee, Wee Sun},
   year={2017}}

@Inbook{pre-image,
author="Kaelbling, Leslie Pack
and Lozano-P{\'e}rez, Tom{\'a}s",
title="Pre-image Backchaining in Belief Space for Mobile Manipulation",
bookTitle="ISRR",
year="2017"
}

@article{realtime_mpc,
  author = {Richards, Arthur and How, J. P}, 
  title = {{Real-Time Model Predictive Control: A Framework for Optimal Guidance and Control in Aerospace Systems}},
  journal = {JGCD},
  year = 2004
}

@article{GBS,
author = {Vadim Indelman and Luca Carlone and Frank Dellaert},
title ={Planning in the continuous domain: A generalized belief space approach for autonomous navigation in unknown environments},
journal = {IJRR},
year = {2015}
}

@article{Kurniawati,
  author       = {Hanna Kurniawati},
  title        = {Partially Observable Markov Decision Processes (POMDPs) and Robotics},
  journal      = {CoRR},
  year         = {2021}
}

@article{likhachev2009planning,
  title={Planning long dynamically feasible maneuvers for autonomous vehicles},
  author={Likhachev, Maxim and Ferguson, Dave},
  journal={IJRR},
  year={2009},
  publisher={SAGE Publications Sage UK: London, England}
}

@inproceedings{TAMPSequencingFirst,
  title={Combined task and motion planning through an extensible planner-independent interface layer},
  author={Srivastava, Siddharth and Fang, Eugene and Riano, Lorenzo and Chitnis, Rohan and Russell, Stuart and Abbeel, Pieter},
  booktitle={ICRA},
  year={2014}
}

@article{TAMPSatisfactionFirst,
  title={Randomized multi-modal motion planning for a humanoid robot manipulation task},
  author={Hauser, Kris and Ng-Thow-Hing, Victor},
  journal={IJRR},
  year={2011},
  publisher={SAGE Publications Sage UK: London, England}
}

@article{TAMPInterleaved,
  title={Semantic attachments for domain-independent planning systems},
  author={Dornhege, Christian and Eyerich, Patrick and Keller, Thomas and Tr{\"u}g, Sebastian and Brenner, Michael and Nebel, Bernhard},
  journal={Towards Service Robots for Everyday Environments},
  year={2012}
}

@article{effectiveBC,
  title={Tossingbot: Learning to throw arbitrary objects with residual physics},
  author={Zeng, Andy and Song, Shuran and Lee, Johnny and Rodriguez, Alberto and Funkhouser, Thomas},
  journal={TRO},
  year={2020}
}

@inproceedings{domainrandomization,
  title={Domain randomization for transferring deep neural networks from simulation to the real world},
  author={Tobin, Josh and Fong, Rachel and Ray, Alex and Schneider, Jonas and Zaremba, Wojciech and Abbeel, Pieter},
  booktitle={IROS},
  year={2017},
}

@article{VLMManipulate,
  title={Manipulate-Anything: Automating Real-World Robots using Vision-Language Models},
  author={Duan, Jiafei and Yuan, Wentao and Pumacay, Wilbert and Wang, Yi Ru and Ehsani, Kiana and Fox, Dieter and Krishna, Ranjay},
  journal={arXiv:2406.18915},
  year={2024}
}

@inproceedings{LLMHighlevelreasoner,
  title={Do as i can, not as i say: Grounding language in robotic affordances},
  author={Brohan, Anthony and Chebotar, Yevgen and Finn, Chelsea and Hausman, Karol and Herzog, Alexander and Ho, Daniel and Ibarz, Julian and Irpan, Alex and Jang, Eric and Julian, Ryan and others},
  booktitle={CoRL},
  year={2023}
}

@book{camacho2013model,
  title={Model Predictive Control},
  author={Camacho, E.F. and Alba, C.B.},
   year={2013}
}

@article{likhachev2013,
author = {Benjamin Cohen and Sachin Chitta and Maxim Likhachev},
title ={Single- and dual-arm motion planning with heuristic search},
journal = {IJRR},
year = {2014},
}

@inproceedings{efficientDiffPolicy,
author = {Kang, Bingyi and Ma, Xiao and Du, Chao and Pang, Tianyu and Yan, Shuicheng},
title = {Efficient diffusion policies for offline reinforcement learning},
year = {2024},
booktitle = {NIPS},
}

@article{steap,
  author={Mukadam, Mustafa and Dong, Jing and Dellaert, Frank and Boots, Byron},
  title={STEAP: Simultaneous Trajectory Estimation and Planning},
  journal = {Autonomous Robots},
  year = 2019
}

@article{OMPL,
    Author = {Ioan A. {\c{S}}ucan and Mark Moll and Lydia E. Kavraki},
    Doi = {},
    Journal = {RAM},
    Month = {},
    Number = {},
    Pages = {},
    Title = {The {O}pen {M}otion {P}lanning {L}ibrary},
    Note = {\url{https://ompl.kavrakilab.org}},
    Volume = {},
    Year = {2012}
}

@INPROCEEDINGS{self-driving,
  author={Jeong hwan Jeon and Cowlagi, Raghvendra V. and Peters, Steven C. and Karaman, Sertac and Frazzoli, Emilio and Tsiotras, Panagiotis and Iagnemma, Karl},
  booktitle={ACC}, 
  title={Optimal motion planning with the half-car dynamical model for autonomous high-speed driving}, 
  year={2013},
  volume={},
  number={},
  pages={},
  doi={10.1109/ACC.2013.6579835}}

@INPROCEEDINGS{SB_manipulation,
  author={Perez, Alejandro and Karaman, Sertac and Shkolnik, Alexander and Frazzoli, Emilio and Teller, Seth and Walter, Matthew R.},
  booktitle={IROS}, 
  title={Asymptotically-optimal path planning for manipulation using incremental sampling-based algorithms}, 
  year={2011},
  volume={},
  number={},
  pages={4307-4313},
  doi={10.1109/IROS.2011.6094994}}

@article{GCS,
   title={Shortest Paths in Graphs of Convex Sets},
   volume={34},
   ISSN={1095-7189},
   DOI={10.1137/22m1523790},
   number={1},
   journal={SIOPT},
   author={Marcucci, Tobia and Umenberger, Jack and Parrilo, Pablo and Tedrake, Russ},
   year={2024},
   month=feb, pages={507–532} 
}

@article{ma2023eureka,
    title   = {{Eureka: Human-Level Reward Design via Coding Large Language Models}},
    author  = {Yecheng Jason Ma and William Liang and Guanzhi Wang and De-An Huang and Osbert Bastani and Dinesh Jayaraman and Yuke Zhu and Linxi Fan and Anima Anandkumar},
    year    = {2023},
    journal = {arxiv-2310.12931}
}

@misc{lstm-gmm,
      title={Generating Sequences With Recurrent Neural Networks}, 
      author={Alex Graves},
      year={2014},
      eprint={1308.0850},
      archivePrefix={arXiv},
      primaryClass={cs.NE}
}

@ARTICLE{linear-mpc,
  author={},
  journal={IEEE Transactions on Automatic Control}, 
  title={Constrained Optimal Control of Linear and Hybrid Systems}, 
  year={2005},
  volume={50},
  number={7},
  pages={1069-1070},
  doi={10.1109/TAC.2005.851468}}

@INPROCEEDINGS{npmpc,
  author={Zheng, A.},
  booktitle={Proceedings of the 1997 American Control Conference (Cat. No.97CH36041)}, 
  title={A computationally efficient nonlinear MPC algorithm}, 
  year={1997},
  volume={3},
  number={},
  pages={1623-1627 vol.3},
}

@article{AO-sbmp,
   author = "Gammell, Jonathan D. and Strub, Marlin P.",
   title = "Asymptotically Optimal Sampling-Based Motion Planning Methods", 
   journal= "Annual Review of Control, Robotics, and Autonomous Systems",
   year = "2021",
   volume = "4",
   number = "Volume 4, 2021",
   pages = "295-318",
  }

@article{rekep,
  title={ReKep: Spatio-Temporal Reasoning of Relational Keypoint Constraints for Robotic Manipulation},
  author={Huang, Wenlong and Wang, Chen and Li, Yunzhu and Zhang, Ruohan and Fei-Fei, Li},
  year={2024}
}

@article{sayplan,
  title={Sayplan: Grounding large language models using 3d scene graphs for scalable task planning},
  author={Rana, Krishan and Haviland, Jesse and Garg, Sourav and Abou-Chakra, Jad and Reid, Ian D and Suenderhauf, Niko},
  journal={CoRR},
  year={2023}
}

@article{safe_learning,
   author = "Brunke, Lukas and Greeff, Melissa and Hall, Adam W. and Yuan, Zhaocong and Zhou, Siqi and Panerati, Jacopo and Schoellig, Angela P.",
   title = "Safe Learning in Robotics: From Learning-Based Control to Safe Reinforcement Learning", 
   journal= "Annual Review of Control, Robotics, and Autonomous Systems",
   year = "2022",
   volume = "5",
   number = "Volume 5, 2022",
   pages = "411-444",
   publisher = "Annual Reviews",
   issn = "2573-5144",
   type = "Journal Article"
  }

\end{document}